\documentclass[11pt]{article}

\usepackage{acl}

\usepackage{times}
\usepackage{latexsym}

\usepackage[T1]{fontenc}

\usepackage[utf8]{inputenc}

\usepackage{microtype}

\usepackage{inconsolata}

\usepackage{graphicx}

\usepackage{amssymb}
\usepackage{amsmath} 
\usepackage{makecell}
\usepackage{multicol}
\usepackage{multirow}
\usepackage{pifont}
\usepackage[table]{xcolor}
\usepackage{colortbl}
\usepackage{subcaption}
\usepackage{graphicx}
\usepackage{booktabs}
\usepackage{wrapfig}
\usepackage{enumitem}
\usepackage{listings}
\usepackage{algorithm}
\usepackage{algpseudocode}

\newcommand{\cmark}{\checkmark}

\usepackage{titlesec}
\titlespacing*{\paragraph}{0pt}{1ex}{0.6em}

\usepackage[most]{tcolorbox}
\newtcolorbox{insightbox}{
  colback=gray!4,
  colframe=gray!30,
  boxrule=0.8pt,
  arc=4pt,
  left=3pt,
  right=3pt,
  top=3pt,
  bottom=3pt,
  width=\linewidth,
  before skip=3pt,
  after skip=1.5pt,
  fontupper=\textbf{\textit{{\fontsize{10pt}{11pt}\selectfont Insight: }}}\itshape
}

\title{LoopMTP: A looped transformer guided by latent multi-token prediction}

\author{
    Behzad Shomali\textsuperscript{1,2} \   
    Markus Frey\textsuperscript{1,2,3} \
    David Berghaus\textsuperscript{1,3} \
    Joachim Koehler\textsuperscript{1,3}  \
    Mehdi Ali\textsuperscript{1,3} \\[6pt]
    \textsuperscript{1}Lamarr Institute \
    \textsuperscript{2}University of Bonn \
    \textsuperscript{3}Fraunhofer IAIS \ \\[6pt]
    \texttt{behzad.shomali@uni-bonn.de}
}

\begin{document}
\maketitle
\begin{abstract}
Looped transformers have emerged as a parameter-efficient alternative to scaling depth for strong reasoning. By reusing one stack of layers across $T$ iterations, they attain the effective depth and reasoning capabilities of larger models at a fixed parameter count. Yet existing approaches suffer from latent overthinking and undifferentiated computation, largely because intermediate representations receive no guidance across loops. Multi-token prediction (MTP) supplies exactly the dense, forward-looking supervision the loop is missing. We propose \textsc{LoopMTP}, which links the two through a structural correspondence in latent space: a model that loops $T$ times can anticipate $T$ future tokens. 
\textsc{LoopMTP} realizes this by softly aligning the hidden state of loop $t$ with the embedding of the token $t$ steps ahead, while a lightweight gate preserves useful information across iterations. \textsc{LoopMTP} improves average accuracy by up to 8.1\% (relative) over the non-looped baseline, with training remaining stable for up to 15 loops.
\end{abstract}

\section{Introduction}

Large language models (LLMs) have demonstrated strong reasoning capabilities that translate into substantial gains on downstream applications such as mathematics and code generation \cite{guo2025deepseek, shao2024deepseekmath}. From an architectural perspective, these capabilities have historically been tied to scale, and in particular to depth. Many reasoning tasks benefit from applying more sequential computation to an input \cite{saunshi2025reasoning}, which standard transformers realize by stacking additional layers. As a result, strong reasoners tend to be very large.
Orthogonal to increasing model size, a separate line of work improves reasoning by modifying the pretraining objective itself. Rather than relying solely on next-token prediction, these approaches adopt richer objectives such as multi-token prediction (MTP), which trains the model to predict the next $k$ tokens simultaneously. By forcing the model to \textit{think ahead}, MTP has been shown to strengthen the reasoning capabilities of LLMs \cite{liu2024deepseek, gloeckle2024better}. However, the benefits of MTP have so far been demonstrated only at larger scale.

Yet access to such models is usually constrained: frontier models are often available only via API, and many organizations 
cannot host
competitive open-weight models locally. In domains such as healthcare, this is a hard requirement; sensitive data cannot leave the premises, so models must run on-site, often on limited hardware \cite{garg2026rise}. Strong reasoning is thus increasingly needed under a parameter-efficient budget.

Looped transformers have emerged as a promising research avenue toward this goal. Rather than adding parameters, they reuse the same stack of transformer blocks across several iterations, a form of \textit{latent reasoning} in which the model refines its representation over several loops before committing to a token. This iterative reuse of computation blocks simulates the effective depth of a much larger model at a fixed parameter count, and has been shown to match the reasoning capabilities of significantly larger models \cite{fu2026simply}.

Despite their success, current looped models suffer from the following limitations. First, they overwrite intermediate computation: each iteration replaces the previous iteration's hidden representation, discarding information that may still be useful \cite{jeddi2026loopformer, zhu2025scaling}. This gives rise to \textit{latent overthinking} \cite{fu2025think}, where predictions that are already correct after an early pass are revised into errors as looping continues. Second, they suffer from \textit{undifferentiated computation}: as the number of iterations grows, hidden representations become increasingly similar across iterations, so successive iterations perform redundant work and waste compute \cite{yu2025mesh}.

To address these limitations, we (1) propose \textsc{LoopMTP}, a novel looped transformer architecture that leverages MTP in latent space as a natural remedy for latent overthinking and undifferentiated computation, (2) study the effect of the MTP alignment objective and the aggregation mechanism on downstream performance, (3) demonstrate that our approach outperforms current looped transformers, and (4) show how our architecture can be used to train small domain-specific expert models.

\section{Related Work} 

\paragraph{Looped transformers \& latent reasoning.} Looped transformers reuse the same set of parameters across multiple forward passes, achieving the computational depth of larger non-looped models without increasing parameter count \citep{dehghani2018universal, zeng2026ponderlm}. Instead of a fixed amount of computation, adaptive looping has also been investigated, where a learned halting mechanism decides how many iterations each input requires \citep{banino2021pondernet, zhu2025scaling, frey2026adaptive}.
Looping in transformers can proceed along two axes both representing a form of latent reasoning. Vertical looping reapplies the same block of layers to iteratively refine the token representations, adding effective depth without lengthening the sequence \citep{saunshi2025reasoning}. 
Horizontal looping instead unfolds across the sequence. A hidden state is fed back as an additional latent token, carrying reasoning into latent space before any discrete token is emitted \citep{hao2024training, shen2025codi}. 
Our approach falls in the vertical category: we loop over the model and aggregate representations across iterations, preserving information rather than overwriting it.

\paragraph{Multi-token prediction.} Next-token prediction provides a single supervision signal per position, which may limit the model's ability to plan ahead \citep{cornille2024learning}. 
Multi-token prediction (MTP) addresses this by training the model to forecast several future tokens at once, yielding denser gradients and encouraging representations that account for longer-range dependencies \citep{liu2024deepseek}.
While MTP has been widely adopted for speculative decoding at inference time, a separate line of work leverages it to improve training itself \citep{gloeckle2024better, liu2024deepseek}. 
With regard to the usage of MTP specifically, the closest work to ours is \citet{noci2026thinking}, who combine MTP with latent reasoning. At selected positions, the model performs a multi-step lookahead in hidden space, supervised against the upcoming ground-truth tokens via a cross-entropy loss over the entire vocabulary.
We reduce this cost by using MTP in latent space as a \textit{soft guiding signal}. 
Rather than predicting a distribution over the vocabulary, we use cosine similarity to align each iteration's hidden state with the embedding of a specific future token. 
Because this signal is applied at every iteration, its cost scales with the number of loops. 
A cross-entropy objective would incur one vocabulary projection per iteration, whereas our cosine alignment adds negligible overhead even at high loop counts. 
Moreover, whereas \citet{noci2026thinking} focus on synthetic planning tasks, we target reasoning in parameter-efficient looped models.

\section{Methodology}
\label{sec:method} 

\begin{figure*}[t]
    \centering
    \includegraphics[
        width=0.92\linewidth,
        trim=0cm 0pt 0cm 8pt, %
        clip
    ]{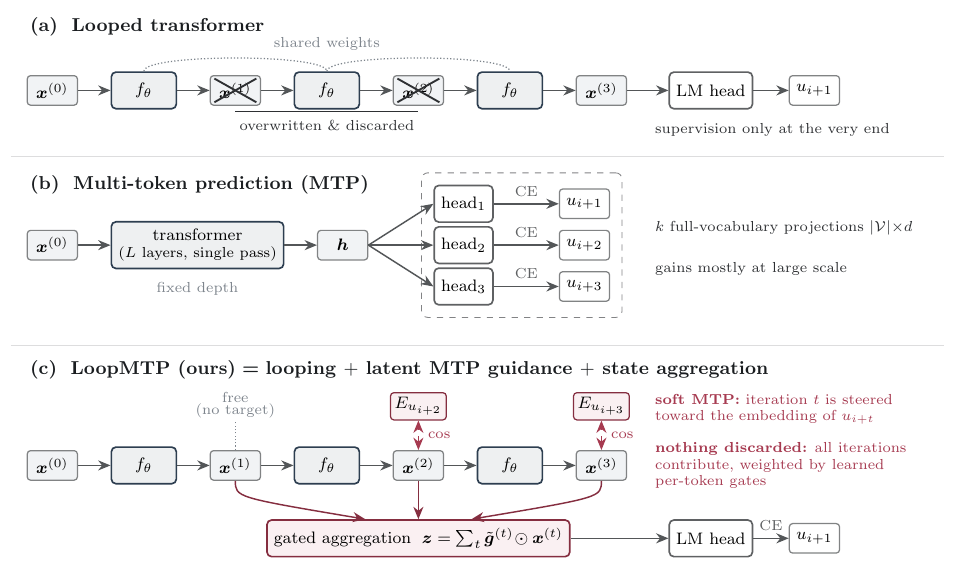}
    \vspace{-8pt}
    \caption{\textbf{Architecture and mechanism overview.} 
    (a)~A standard looped transformer block applied $T$ times where next-token supervision is applied only at the very end.
    (b)~A standard multi-token prediction (MTP) model requiring full-vocabulary projections.
    (c)~\textbf{\textsc{LoopMTP}} (ours): per-iteration outputs are aligned in the latent space with future token embeddings (soft MTP) and combined via a gating mechanism. Illustrated for a model with $T{=}3$ loops.}
    \vspace{-7pt}
    \label{fig:LoopMTP_arch}
\end{figure*}

Figure~\ref{fig:LoopMTP_arch} contrasts our proposed method, \textsc{LoopMTP}, with standard looped transformers and multi-token prediction. \textsc{LoopMTP} addresses representation overwriting and latent overthinking, and undifferentiated computation through an MTP-guided looped block (Section~\ref{sec:method:cond}), learned state aggregation (Section~\ref{sec:method:agg}), and a soft-alignment auxiliary objective (Section~\ref{sec:method:loss}).

\subsection{Notation}
In the following, we build upon the notation of \citet{saunshi2025reasoning}. Let $f_\theta$ denote a stack of $L$ transformer blocks with shared
parameters $\theta$, and $T$ denote the number of times this stack is applied.
Given an input token sequence $\mathbf{u} {=} (u_1, \dots, u_S) \in \mathcal{V}^S$, where $\mathcal{V}$ denotes the vocabulary and $S$ the sequence length, we first embed the tokens:
\setlength\abovedisplayskip{5pt}
\setlength\belowdisplayskip{6pt}
\begin{equation}
  \mathbf{x}^{(0)} \;=\; \mathrm{Embed}(\mathbf{u}) \in \mathbb{R}^{S \times d},
\end{equation}
and then apply $f_\theta$ recursively for $t = 1, \dots, T$:
\begin{equation}
  \mathbf{x}^{(t)} \;=\; f_\theta\!\left(\mathbf{x}^{(t-1)}, \mathbf{x}^{(0)}\,,\, t\right).
  \label{eq:loop}
\end{equation}
The explicit dependence on $\mathbf{x}^{(0)}$ and the iteration index $t$
is discussed in Section~\ref{sec:method:cond}.

\subsection{MTP-Guided Looped Block}
\label{sec:method:cond}
Our MTP-guided looped block $f_{\theta}$ comprises all $L$ transformer blocks, i.e., we loop over the entire model.
Each application of $f_\theta$ at iteration $t$ takes as input the previous
iteration's output $\mathbf{x}^{(t-1)}$ and the fixed token embeddings
$\mathbf{x}^{(0)}$, and outputs $\mathbf{x}^{(t)}$. We construct the
input to $f_\theta$ via three operations: (i) a normalized iteration-index
embedding, (ii) per-iteration normalization, and (iii) a concatenation-and-projection step.

\paragraph{Iteration-index embedding and per-iteration normalization.}
Following \citet{frey2026adaptive}, we explicitly signal the current iteration by appending the normalized iteration index to token vectors.
We then normalize hidden states with iteration-specific LayerNorms $\mathrm{LN}^{\text{prev}}_{t}$ and obtain token embeddings using
$\mathrm{LN}^{\text{tok}}$:
\begin{equation}
  \mathbf{h}^{(t-1)} \;{=}\; \mathrm{LN}^{\text{prev}}_{t}\!\left(\big[\, \frac{t{-}1}{T} \;\Vert\; \mathbf{x}^{(t-1)} \,\big]\right) {\in } \mathbb{R}^{S \times (d+1)}, 
\end{equation}
\begin{equation}
  \mathbf{e}   \;=\; \mathrm{LN}^{\text{tok}}\!\left(\mathbf{x}^{(0)}\right) \in \mathbb{R}^{S \times d},
\end{equation}
where $\Vert$ denotes feature-axis concatenation.
Per-iteration norms are crucial, letting each loop operate on inputs with iteration-specific statistics.

\paragraph{Input fusion.}
The previous-iteration state and the re-normalized token embedding are merged
by concatenation along the feature axis followed by a linear projection $\mathrm{P}$:

\begin{equation}
  \mathbf{y}^{(t)} \;=\; \mathrm{P}\!\left(\big[\, \mathbf{e} \;\Vert\; \mathbf{h}^{(t-1)} \,\big]\right) \in \mathbb{R}^{S \times d}.
\end{equation}

The shared backbone is then applied:
\begin{equation}
  \mathbf{x}^{(t)} \;=\; f_\theta\!\left(\mathbf{y}^{(t)}\right)
                    \;=\; \mathrm{Layer}_L \circ \cdots \circ \mathrm{Layer}_1\!\left(\mathbf{y}^{(t)}\right).
\end{equation}

\paragraph{Looped layer-norm scaling (Loop-LNS).}
To preserve stability under a variable number of unrollings, we adapt layer-norm scaling (LNS) \citep{sun2026curse} to the looped setting. Inside each layer $\ell$, the input $\mathbf{x}$ is first normalized (i.e.~pre-norm transformer) and rescaled by a fixed factor $\frac{1}{T}$:
\begin{align}
  \mathbf{x} \;=\; \mathbf{x} + \mathrm{Attn}\!\left(\tfrac{1}{T}\,\mathrm{RMSNorm}(\mathbf{x})\right),
  \\
  \mathbf{x} \;=\; \mathbf{x} + \mathrm{FFN}\!\left(\tfrac{1}{T}\,\mathrm{RMSNorm}(\mathbf{x})\right).
\end{align}
Unlike standard LNS \citep{sun2026curse}, which scales by $\frac{1}{\sqrt{n}}$, where $n$ is the running depth, our \emph{Loop-LNS} variant uses a fixed factor of $\frac{1}{T}$ for all 
effective sub-blocks, reflecting the fact that the backbone is reused $T$ times rather than deepened.

We chose the fixed per-iteration factor $\frac{1}{T}$ over running-depth alternatives for a specific reason: a progressive counter that scales layer~$\ell$ in iteration~$t$ by $\frac{1}{(t{-}1) \cdot L + \ell}$  yields tiny factors for later iterations of deep unrollings, which we found empirically neutralizes their gradient contribution. 
The constant $\frac{1}{T}$ balances residual-stream stability with sufficient signal propagation to all iterations.

\paragraph{Per-Iteration Representation Alignment with MTP. }
To exploit the per-iteration hidden states $\{\mathbf{x}^{(t)}\}_{t=2}^{T}$, we encourage each iteration's hidden state to point toward the output embedding of the token it should anticipate. 
For iteration $t {\geq} 2$, position $i$ is aligned to the output embedding of token $u_{i+t}$, so that the $t$-th iteration is responsible for anticipating the token $t$ steps ahead.
The first iteration $t{=}1$ is reserved as an unconstrained representation (i.e.~no supervision): the model is free to learn what kind of information to encode here, so that this representation is rich enough for subsequent iterations to read out multiple future tokens from it.

\subsection{Aggregation}
\label{sec:method:agg}

Rather than discarding the intermediate iterations, we aggregate
$\{\mathbf{x}^{(t)}\}_{t=1}^{T}$ via a token-wise weighted sum with a
\emph{shared, content-conditional} gate. %
Let $W_g \in \mathbb{R}^{d \times d}$
be a single linear gate shared across all iterations and let
$\boldsymbol{\beta} = (\beta_1, \dots, \beta_T) \in \mathbb{R}^{T}$
be per-iteration scalar bias parameters. For each iteration $t$ and position
$i \in \{1,\dots,S\}$, we compute an unnormalized gate:
\begin{equation}
  g_i^{(t)} = \text{softplus}\!\Big(W_g\, \mathbf{x}_i^{(t)} + \beta_t \cdot \mathbf{1}_d\Big)
  \;\in\; \mathbb{R}^d,
\end{equation}
where $\mathbf{1}_d \in \mathbb{R}^d$ is the all-ones vector. We then normalize gate values across iterations:
\begin{equation}
  \tilde{\mathbf{g}}^{(t)}_{i} {=} \frac{\mathbf{g}^{(t)}_{i}}{\sum_{s=1}^{T} \mathbf{g}^{(s)}_{i} + \varepsilon},
\end{equation}
where $\varepsilon = 10^{-8}$ to avoid division by zero. 
Finally, the aggregation happens as follows:
\begin{equation}
    \mathbf{z}_{i} \;=\; \sum_{t=1}^{T} \tilde{\mathbf{g}}^{(t)}_{i} \odot \mathbf{x}^{(t)}_{i},
  \label{eq:ws}
\end{equation}
where $\odot$ is the elementwise product. 
The gating mechanism itself introduces a linear layer of size $d {\times} d$ and $T$ learnable scalar bias terms, amounting to ${<}0.5\%$ additional parameters for our models.

Similarly to \citet{frey2026adaptive}, we found the gate initialization important: 
the exact initial value matters less than how the model behaves at the beginning of training.
The first bias term is initialized to a moderately positive value, while all subsequent bias terms start at a substantially negative value. This provides a strong prior toward the first iteration, i.e., acting as a vanilla non-looped transformer at first, while leaving later iterations free to take over as training progresses.

\subsection{Training Objective}
\label{sec:method:loss}

The total loss combines three terms: the main next-token cross-entropy, a hidden-state alignment loss, and a
ponder regularizer.

\paragraph{Main NTP loss.} The next-token prediction (NTP) loss, commonly used as the main pretraining objective for LLMs, is applied to the output of the language-model head on the normalized aggregated representation $\mathbf{z}$ (denoted by $\mathbf{p}_i^{\text{agg}}$):
\begin{equation}
  \mathcal{L}_{\text{NTP}} \;=\; -\frac{1}{S-1}\sum_{i=1}^{S-1} \log \mathbf{p}_i^{\text{agg}}\!\left[u_{i+1}\right].
\end{equation}

\paragraph{Hidden-state soft MTP alignment.}
We regularize the per-iteration hidden states to align with the embeddings of the tokens they are tasked with predicting. Let $E \in \mathbb{R}^{|\mathcal{V}| \times d}$ denote the output (unembedding) matrix, treated as a fixed target via stop-gradient $\mathrm{sg}[\cdot]$. For $t = 2, \dots, T$, the per-step hidden-state alignment loss is:
\begin{equation}
  \mathcal{L}^{(t)}_{\text{align}} \;=\; \frac{1}{S - t}\sum_{i=1}^{S - t}\!\left(1 - \cos\!\big(\mathbf{x}^{(t)}_{i},\, \mathrm{sg}[E_{u_{i+t}}]\big)\right).
\end{equation}
The final alignment loss averages these:
\begin{equation}
  \mathcal{L}_{\text{align}} \;=\; \frac{1}{T-1}\sum_{t=2}^{T}\mathcal{L}^{(t)}_{\text{align}}.
\end{equation}

Notably, in this case we align the per-iteration \emph{hidden state} with the future token \emph{embedding}, in contrast to $\mathcal{L}_\text{NTP}$, where the alignment is over the \emph{distribution}. By matching embeddings rather than full output distributions, our objective imposes a softer constraint, which is why we refer to it as \emph{soft} MTP alignment. 
We emphasize that our alignment approach does not predict multiple future tokens
in the traditional sense: each iteration's hidden state is steered toward a future
token's embedding via cosine similarity, acting as a \emph{representational regularizer}
rather than a distributional constraint.  We retain the term ``multi-token
prediction'' to reflect the lookahead structure of the supervisory signal, while
noting that this cosine-based formulation avoids the vocabulary-sized projection that makes standard MTP expensive---a property that we believe contributes to its
effectiveness at small scale.
Importantly, computing $\mathcal{L}_\text{align}$ requires only a cosine similarity between each iteration's output and its corresponding target, adding negligible overhead.

\paragraph{Ponder regularizer.}
The aggregator's gates induce a per-token distribution
$G_i(t) {=} \tfrac{1}{d}\sum_{k} \tilde{\mathbf{g}}^{(t)}_{i,k}$ over the
$T$ iterations. 
Following \citet{zhu2025scaling}, we pull this distribution toward the uniform
prior $Q {=} (1/T, \dots, 1/T)$ over the $T$
iterations via the Kullback--Leibler (KL) divergence:
\begin{equation}
  \mathcal{L}_{\text{ponder}} \;=\; \frac{1}{S}\sum_{i=1}^{S} \mathrm{KL}\!\left(G_i \,\Vert\, Q\right).
\end{equation}

\paragraph{Final loss.}
The final loss is calculated as the weighted sum of next token prediction loss, alignment loss, and the ponder loss as follows:
\begin{equation}    
    \mathcal{L} = \mathcal{L}_{\text{NTP}}
      + \lambda_{\text{align}}\, \mathcal{L}_{\text{align}}
      + \lambda_{\text{ponder}}\, \mathcal{L}_{\text{ponder}}
  \label{eq:total}
\end{equation}

\section{Results}
\label{sec:results}

\subsection{Experimental Setup}

\paragraph{Models.} 

The \textsc{LoopMTP} model is a GPT-2-style decoder-only transformer. In our experiments, we set the number of layers to $L=12$, the embedding dimension to $d=1024$, the number of attention heads to 32, and the nominal FFN hidden dimension to 4096. We employ rotary positional embeddings \citep{su2024roformer}, apply RMSNorm to queries and keys prior to attention \citep{henry2020query, dehghani2023scaling}, and use SwiGLU \citep{shazeer2020glu} in the FFNs. 
We compare \textsc{LoopMTP} against two baselines: (1) a non-looped baseline, and (2) LoopFormer \citep{jeddi2026loopformer}, the current state-of-the-art among looped models. The non-looped baseline keeps the same architectural hyperparameters as \textsc{LoopMTP}, except that its nominal FFN hidden dimension is increased to 4352 (realized width 3072), which more than compensates for the parameters introduced by our gating module and input-fusion projection, giving the baseline roughly 6M more parameters than \textsc{LoopMTP}.
At the same depth and width, LoopFormer has approximately 20M more parameters than our model.

\paragraph{Training.}

All models are trained on the \textit{high-quality} subset of \texttt{Nemotron-CC-v2} \citep{nvidia2025nvidianemotronnano2} and \texttt{Nemotron-CC-Math-v1} \citep{mahabadi2025nemotron} for 6.8B tokens. The models are optimized using the Muon \citep{liu2025muon} and AdamW \citep{loshchilov2017decoupled} optimizers with a peak learning rate of $1.9\times10^{-3}$, a cosine warmup over $0.001$ of total training steps, and a cosine decay schedule down to $1/10$ of the peak learning rate. We train LoopFormer on the same data. Because LoopFormer training was highly unstable across loop configurations, we ran a grid search over learning rate, weight decay, and warmup ratio, yielding 12 configurations per loop count, and report the best result for each.
Appendices~\ref{sec:appendix:implementation_details},~\ref{sec:appendix:WD}, and ~\ref{sec:flops}
provide further details on training and implementation for both approaches, the role of weight decay, and a FLOPs comparison.

\paragraph{Evaluation.}

In our experiments, we report perplexity on a random 5M-token subset of FineWeb-Edu \citep{lozhkov2024fineweb-edu} and OpenWebText \citep{Gokaslan2019OpenWeb}. Downstream performance is evaluated on three groups of tasks using the OLMES framework~\citep{gu2025olmes}: (1) commonsense benchmarks: ARC-Challenge (ARC-C) and ARC-Easy (ARC-E) \citep{allenai:arc}, HellaSwag (HS) \citep{zellers2019hellaswag}, LAMBADA (LB) \citep{paperno2016lambada}, PIQA \citep{bisk2020piqa}, SocialIQA (SIQA) \citep{sap2019social}, and Winogrande (WG) \citep{sakaguchi2021winogrande}; (2) a math suite: Algebra, Counting \& Probability, Geometry, Intermediate Algebra, Number Theory, Prealgebra, and Precalculus; and (3) code benchmarks: CodeX HumanEval \citep{chen2021evaluating} and MBPP \citep{austin2021program}. The reported accuracy and BPB values are averaged over 3 random seeds.
Because our experiments are conducted at a small scale, we use bits-per-byte (BPB) as our primary metric, as it is a reliable proxy for downstream performance at larger scales \citep{gadre2025language, heineman2026signal}. 
For benchmarks where accuracy provides a meaningful signal, we report accuracy as well. In Section~\ref{sec:expert}, we report accuracy on GSM8K~\cite{cobbe2021gsm8k} and BPB on the math suite.

\definecolor{lightblue}{RGB}{220,230,255}
\definecolor{lightgray}{RGB}{245,245,245}

\begin{table*}[t]
    \centering
    \resizebox{0.994\linewidth}{!}{
    \begin{tabular}{
        @{}l
        c |
        c c |
        c c c c c c c c |
        c c c c@{}
        }
    \toprule
        & & \multicolumn{2}{c}{\textit{Lang.\ Modeling (PPL $\downarrow$)}} & \multicolumn{8}{c}{\textit{General Tasks (Accuracy $\uparrow$)}} & \multicolumn{4}{c}{\textit{Math/Code/QA (BPB $\downarrow$)}} \\
        \cmidrule(lr){3-4} \cmidrule(lr){5-12} \cmidrule(lr){13-16}
        Model & \# Params & FineWeb-Edu & OpenWebText & ARC-C & ARC-E & HS & WG & SIQA & PIQA & LB & \textit{Avg} & QA & Math & Code & \textit{Avg} \\
        \midrule
        \textbf{Non-looped}          & 266M & 21.08 & 23.56 &  31.60 & 59.90 & 38.48 & 50.96 & 44.85 & 65.98 & 32.16 & 46.28 & 0.9948 & 0.7114 & 0.8624 & 0.8562\\
        
        \midrule
        \textit{\textbf{LoopFormer$^*$ }
        } & \multirow{5}{*}{280M} &&&&&&&&&&&&&&\\
        \quad Loops=3$^\dagger$  & &  21.18 & 23.91        & 32.04 & 58.69 & 38.61 & 50.24 & 44.19 & 66.00 & 30.98 & 45.82 & 1.0212 & 0.7344 & 0.9037 & 0.8864\\
        \quad Loops=5  & &  21.05 & 23.65        & 35.24 & 59.65 & 41.61 & 52.35 & 45.87 & 65.29 & 35.13 & 47.88 & 0.9747 & 0.6898 & 0.8348 & 0.8331\\
        \quad Loops=7  & &  20.66 & 23.20        & 30.01 & 46.49 & 34.37 & 51.43 & 42.39 & 60.52 & 20.84 & 40.86 & 1.5140 & 1.7259 & 2.3012 & 1.8470\\
        \quad Loops=9  & &  20.72 & 23.31        & 32.17 & 49.97 & 36.50 & 50.86 & 42.94 & 60.97 & 23.73 & 42.45 & 1.4555 & 1.6034 & 1.8089 & 1.6226\\

        \midrule
        \rowcolor{lightgray}
        \textit{\textbf{\textsc{LoopMTP}}\small(ours)\normalsize} & \multirow{5}{*}{} &&&&&&&&&&&&&&\\
        \rowcolor{lightgray}
        \quad Loops=3     &   & 19.43 & 21.57 & \underline{35.49} & 62.28 & 42.60 & 50.07 & 46.37 & 67.17 & 35.48 & 48.49 & 0.9541 & 0.6681 & 0.8071 & 0.8098\\
        \rowcolor{lightgray}
        \quad Loops=5 & \multirow{-1}{*}{260M}   & 18.89 & 20.97 & 35.32 & 63.38 & 44.35 & 52.54 & 46.88 & 67.52 & \underline{36.79} & 49.54 & 0.9413 & \underline{0.6584} & \underline{0.7921} & \underline{0.7973}\\
        \rowcolor{lightgray}
        \quad Loops=7     &   & \underline{18.87} & \underline{20.94} & 35.10 & \textbf{64.07} & \textbf{44.78} & \textbf{52.99} & \underline{46.95} & \textbf{67.97} & 36.39 & \underline{49.75} & \underline{0.9369} & \textbf{0.6575} & \textbf{0.7822} & \textbf{0.7922}\\
        \rowcolor{lightgray}
        \quad Loops=9     &  & \textbf{18.86} & \textbf{20.92} & \textbf{36.01} & \underline{63.71} & \underline{44.64} & \underline{52.70} & \textbf{47.78} & \underline{67.86} & \textbf{37.47} & \textbf{50.02} & \textbf{0.9366} & 0.6586 & 0.8003 & 0.7985\\
    \bottomrule
    \end{tabular}
    }
    \vspace{-5pt}
    \caption{\textbf{Main results.} We report perplexity for language modeling tasks, accuracy for general tasks, and BPB for QA, math, and code suites. QA reports BPB on the same seven benchmarks for which accuracy is shown under \textit{General Tasks}.
    $^*$ for LoopFormer \cite{jeddi2026loopformer}, stable training required a separate grid search over learning rate, weight decay, and warmup ratio at each loop count. $^\dagger$ indicates runs whose reported averages are over two random seeds, as one of the three runs diverged.
    }
    \vspace{-10pt}
    \label{tab:main_results}
\end{table*}

\subsection{Downstream Tasks Performance}
\label{sec:main_results}

Table~\ref{tab:main_results} reports model performance on the perplexity benchmarks and downstream tasks. 
The numbers reported in the main experiment are averaged across three random seeds. Detailed results are presented in Appendix~\ref{sec:appendix:multiseed_results}.

Despite being slightly smaller, \textsc{LoopMTP} outperforms the non-looped baseline on nearly every benchmark: not only on reasoning-based tasks,
but also on perplexity, general tasks, and QA suites. On general tasks, it improves average accuracy by up to 8.08\% over the non-looped baseline (50.02\% vs.\ 46.28\%) and 21.76\% over LoopFormer (49.75\% vs.\ 40.86\%). On QA, math, and code tasks, it reduces BPB by up to 7.5\% and 57\% relative to the two baselines (0.7922 vs.\ 0.8562 and 1.8470) and improves perplexity by 10.5\% on FineWeb-Edu and 11.2\% on OpenWebText.
Additionally, at matched loop counts, \textsc{LoopMTP} outperforms LoopFormer in 27 of 28 cases (4 loop counts $\times$ 7 benchmarks), and also achieves better perplexity and QA/math/code performance, without per-loop tuning of stability-critical hyperparameters (learning rate, weight decay, warmup). Only $\lambda_\text{align}$ (which controls the strength of the auxiliary MTP signal) is tuned, and all swept values train stably.
    
\paragraph{\textsc{LoopMTP} scales stably with loop count.}
While the number of unique parameters stays fixed, which governs a looped model's capacity to store factual information \citep{frey2026dual}, perplexity for \textsc{LoopMTP} continues to improve with additional loops.
This is consistent with \citet{zhu2025scaling}, who find that looped models can be stronger at knowledge manipulation.
Perplexity and general-task accuracy improve monotonically in loop count ($T$), whereas QA, math, and code BPB improve through $T{=}7$ and regress slightly at $T{=}9$. Crucially, \textsc{LoopMTP}'s training remains stable, unlike LoopFormer's: one reason for LoopFormer's relatively worse performance is its unstable training across different loop numbers and even random seeds (see Appendix~\ref{sec:appendix:multiseed_results}). 
We leave the analysis of looping limits and upper bounds to future work.

\subsection{MTP Auxiliary Objective Impact}
\begin{figure}[t!]
    \centering
    \includegraphics[
        width=0.98\linewidth,
        trim=0cm 0pt 0cm 7pt, %
        clip
    ]{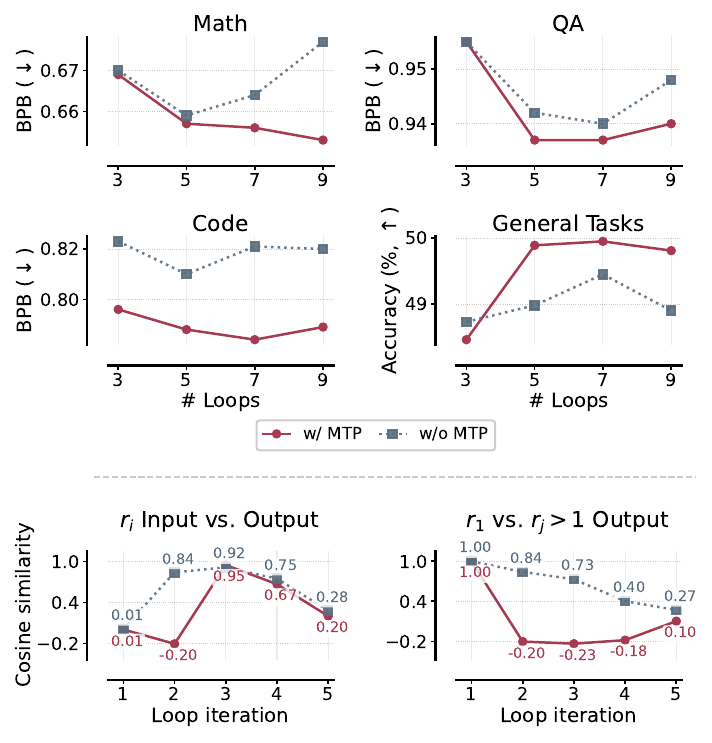}
    \vspace{-10pt}
    \caption{\textbf{Impact of MTP auxiliary training signal on performance and per-loop representations.} (Top) Aligning per-loop representations with the MTP signal improves performance, particularly at higher loop counts. (Bottom) With MTP, at most iterations yield a more distinct representation.
    }
    \vspace{-10pt}
    \label{fig:mtp_effect_performance}
\end{figure}

\paragraph{Effect of MTP on performance.}
Figure~\ref{fig:mtp_effect_performance} (Top) shows the impact of the MTP auxiliary objective on looped transformer performance across loop counts. On the QA/math/code suites, runs with $\lambda_\text{align} {>} 0$ (\textit{w/ MTP}) show clear improvements over runs with $\lambda_\text{align} {=} 0$ (\textit{w/o MTP}). The effect is most pronounced on math, where step-by-step reasoning is required: without MTP, increasing the loop count causes performance to degrade monotonically. General-task average accuracy is also higher with MTP. Together, these results confirm that MTP is an effective guidance signal for looped models, with negligible overhead. Furthermore, the overall results suggest that combining MTP with looped models is a promising direction: unlike \citet{gloeckle2024better}, where MTP hurt performance at a small scale, here it consistently improves results.

\paragraph{Effect of MTP on hidden representations.}
Figure~\ref{fig:mtp_effect_performance} (Bottom) reports cosine similarities for $T{=}5$ models trained with and without the MTP signal: between the input and output of each iteration $r_i$ (left), and between the output of the first iteration and the outputs of subsequent iterations (right). The right panel shows that, for \textit{w/ MTP} model, the second iteration already maps the input to a markedly distinct subspace, and subsequent iterations continue to build on this trajectory. In contrast, the \textit{w/o MTP} model changes representations only gradually across iterations, with similarities remaining relatively high, particularly across the first three iterations. The left panel shows a consistent pattern: at most iterations, the \textit{w/ MTP} model produces a more distinct representation than its non-MTP counterpart,
increasing expressiveness.
These results confirm the existence of \emph{undifferentiated computation}~\citep{yu2025mesh}, visible in the consistently high cross-iteration similarities of \textit{w/o MTP} model, and show that MTP encourages each loop to learn a more unique transformation.

\begin{figure}[t]
    \centering
    \includegraphics[
        width=0.94\linewidth,
        trim=0cm 0pt 0cm 7pt, %
        clip
    ]{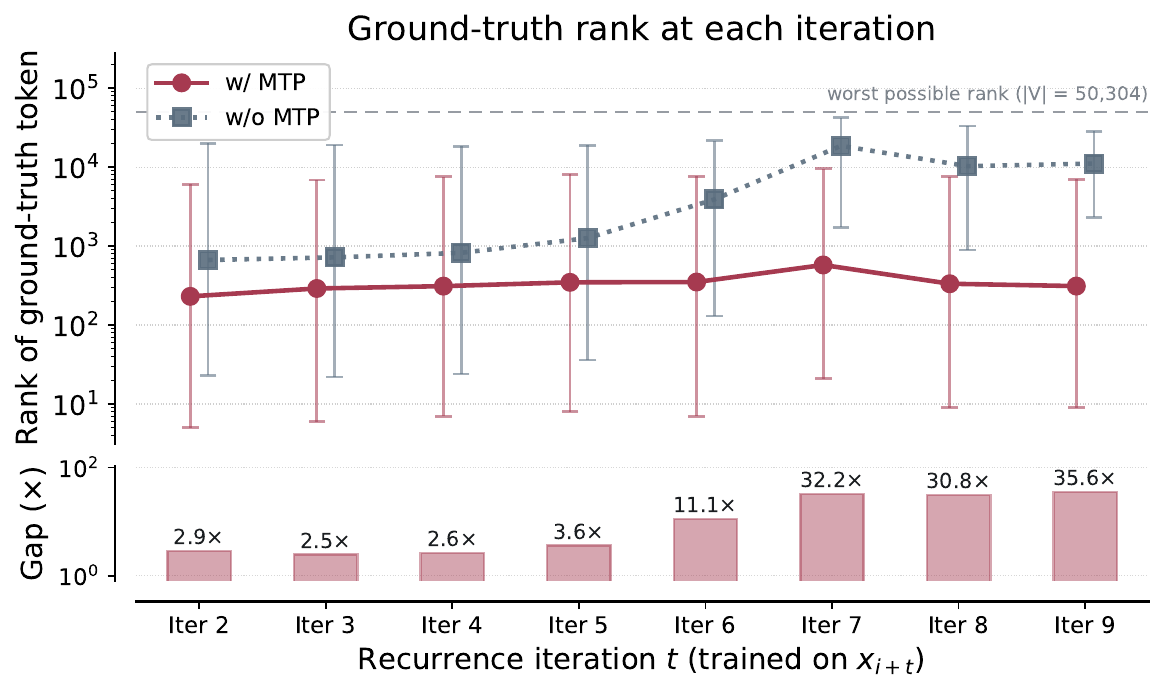}
    \vspace{-10pt}
    \caption{\textbf{MTP training sharpens ground-truth retrieval.} Despite the lightweight nature of the latent MTP guidance, the \textit{w/ MTP} model achieves a ground-truth rank up to 35.6$\times$ better than the \textit{w/o MTP} model. 
    }
    \vspace{-10pt}
    \label{fig:MTP_rank}
\end{figure}

Figure~\ref{fig:MTP_rank} shows the median ground-truth rank (log scale) at each recurrence iteration $t$, evaluated at its trained offset $u_{i+t}$ over 5M tokens from the OpenWebText dataset. This was obtained by feeding the output of each iteration to the language model head and reading off the logit distribution, for a model trained with MTP (\textit{w/ MTP}) and its counterpart trained without (\textit{w/o MTP}). A model that predicts the ground-truth token perfectly assigns it rank one; the lower the rank, the better the model predicts the ground-truth label. The pattern is clear: the rank of the \textit{w/o MTP} model is up to more than an order of magnitude worse than that of the \textit{w/ MTP} model. The \textit{w/ MTP} model, by contrast, stays within a similar range across all future-token predictions. We attribute this to the absence of a discounting factor in our objective: all future tokens are weighted equally.
This confirms the alignment is actually achieved and that it transfers to the language-model head's coordinate system, rather than being satisfied in a subspace the head ignores.

\subsection{Gating Mechanism Impact}

\begin{figure*}[t!]
    \centering
    \includegraphics[
        width=0.91\linewidth,
        trim=0cm 0pt 0cm 11pt, %
        clip
    ]{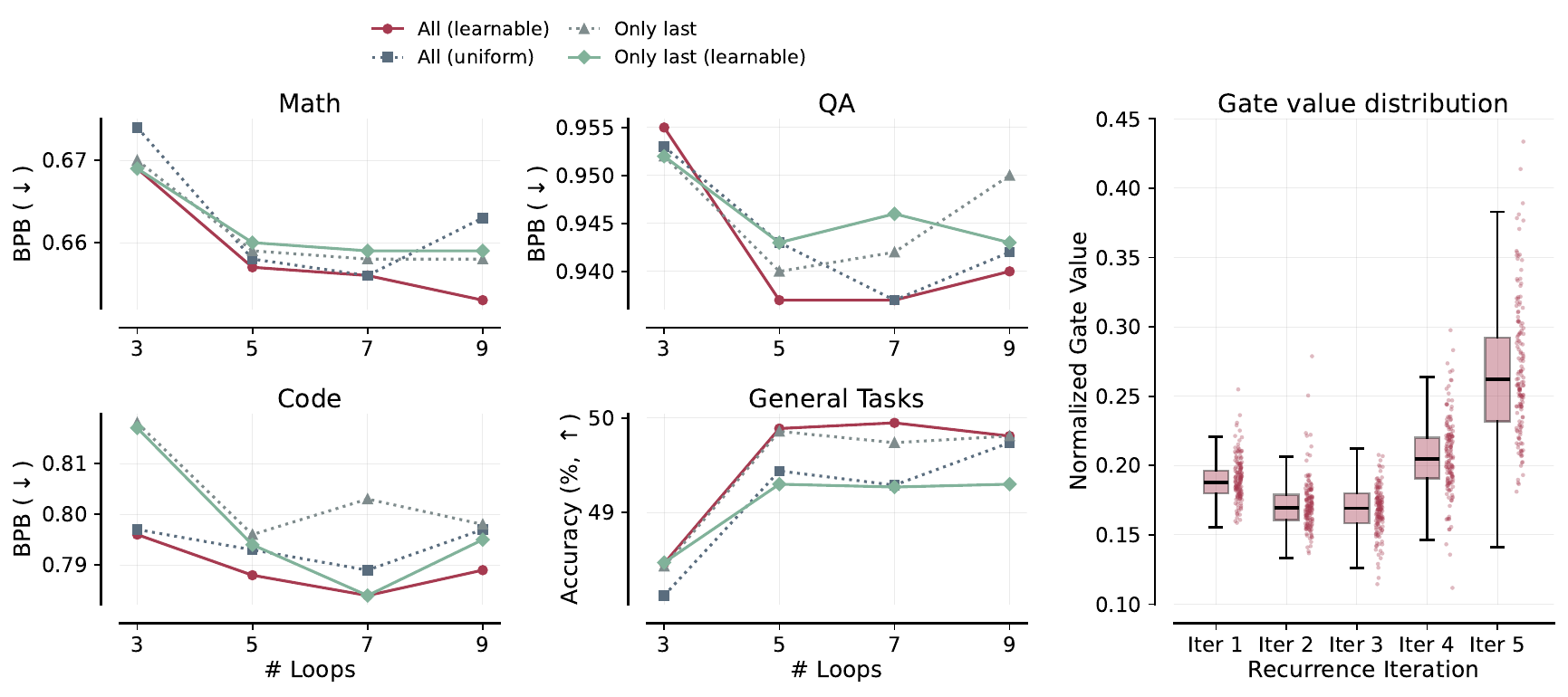}
    \vspace{-11pt}
    \caption{\textbf{Gating mechanism effects.} (Left) When \textit{all gates are learnable} we consistently achieve the best performance, particularly as the number of loops increases.
    (Right) Later iterations exhibit higher and more variable gate values, indicating increasingly input-dependent blending.
    }
    \label{fig:gate_distribution}
\end{figure*}

\paragraph{Gates effectiveness.} Figure~\ref{fig:gate_distribution} (Left) illustrates the importance of the gating mechanism. We compare our per-iteration gating, \textit{All (learnable)}, against three baselines. 
In \textit{Only last}, each iteration overwrites the previous one, with the final gate value fixed to one and all others set to zero. \textit{Only last (learnable)} similarly zeros all non-final gates but allows the final gate value to be determined by a context-dependent learnable mechanism. \textit{All (uniform)} combines the representations by simply taking an average.
\textit{All (learnable)} consistently outperforms all other variants. While general tasks accuracy for \textit{Only last}, \textit{All (uniform)}, and \textit{All (learnable)} is nearly identical, BPB reveals more nuance. Math BPB highlights the importance of representation aggregation, particularly as the number of loops increases: while all other variants either considerably diverge or stagnate, \textit{All (learnable)} continues to decrease BPB with more loops. For QA BPB, the limitation of \textit{Only last} is clear: more iterations means more overwriting and greater information loss. This motivates a mechanism to retain information across loops: even one as simple as averaging, \textit{All (uniform)}, can perform well at higher loop counts. 
This pattern holds for Code as well, where our gating consistently outperforms all variants across loop counts.

\paragraph{Gate behaviour.}
Figure~\ref{fig:gate_distribution} (Right) shows the distribution of normalized gate values for a model with $T{=}5$ across recurrence iterations, evaluated on 5 million randomly sampled tokens from FineWebEdu.
Two trends emerge. First, the mean gate value increases with iteration depth: iterations 1--3 maintain a relatively low and stable mean (${\approx}\,0.17$--$0.19$), whereas iterations 4 and 5 rise to ${\approx}\,0.21$ and ${\approx}\,0.26$, respectively.
This suggests that the gating mechanism learns an \emph{iteration-aware} strategy: early gates act conservatively, yielding a stable scaling value, while later loops integrate newly computed information more dependent on the context.
Second, the spread of the distribution grows markedly in later iterations.
Early iterations exhibit tight, unimodal distributions, indicating that the gate behaves uniformly across tokens.
By contrast, iterations 4 and~5 display substantially wider distributions, implying that the gate adopts a more \emph{input-dependent} policy at greater depth; selectively deciding, on a per-token basis, how much iterative information to preserve.
Taken together, these patterns confirm that the gating mechanism is not merely a static interpolation but learns a structured, depth-dependent blending strategy that enables \textsc{LoopMTP} to benefit from additional recurrence steps without destabilizing training.

\section{Small Domain-Specific Expert}
\label{sec:expert}

In this section, we show that, when trained on a specific domain, \textsc{LoopMTP} can be regarded as a small expert model, well suited to domain-specific reasoning or deployment on memory-constrained hardware. We train models on ${\sim}6.8$B tokens from \texttt{Nemotron-CC-Math-v1} dataset on variable number of loops $T\in \{3,7,9,11,13,15\}$.

\begin{figure}[t!]
    \centering
    \includegraphics[
        width=0.98\columnwidth,
        trim=0cm 0pt 0cm 7pt, %
        clip
    ]{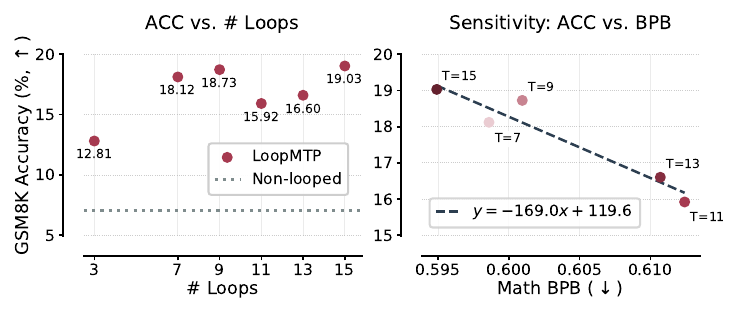}
    \vspace{-10pt}
    \caption{\textbf{GSM8K accuracy for expert \textsc{LoopMTP} and BPB–accuracy sensitivity.} (Left) \textsc{LoopMTP} achieves 19.03\% vs.\ 7.05\% at equal parameter count. (Right) The steep slope shows that even small BPB reductions yield meaningful accuracy gains.
    }
    \vspace{-10pt}
    \label{fig:expert_acc_perplexity}
\end{figure}

Figure~\ref{fig:expert_acc_perplexity} (Left) shows GSM8K accuracy (8-shot) for models trained with varying numbers of loops. Despite the small model size ($\sim$260M parameters) and limited pretraining budget, and without any finetuning or instruction tuning, our model reaches 19.03\% accuracy on GSM8K. At matched parameter count, the non-looped baseline achieves only 7.05\%; a relative improvement of approximately 170\% under the same parameter memory budget. 

One reading of this gap comes from mechanistic work showing that mathematical ability in LLMs is distributed across a localized parameter subset, with performance scaling with the fraction of that subset removed rather than hinging on a few individual weights \citep{shomali2026llm}. What matters for a math expert may therefore be how much of the reused capacity is allocated to math-relevant computation, rather than the raw parameter count. This is precisely what looping provides: more effective capacity at fixed parameter memory.

Accuracy does degrade at certain loop counts: the number of \emph{hurt} samples (initially correct, flipped to incorrect by additional loops) eventually outweighs the number of \emph{rescued} samples at those loop counts (we suspect this reflects the model's underlying reasoning characteristics; Appendix~\ref{sec:appen:mechanistic_verification} reports related token-length statistics).
Figure~\ref{fig:expert_acc_perplexity} (Right) plots GSM8K accuracy against BPB on the math suite across varying numbers of loops, together with a fitted line. The slope of $-169$ indicates that even a small reduction in BPB translates to considerable accuracy gain. For example, a 0.01 reduction in BPB would lead to roughly a 1.69\% increase in accuracy.

\section{Conclusion}

We introduced \textsc{LoopMTP}, which targets two weaknesses of looped models, latent overthinking and undifferentiated computation, with a MTP-guided looped block, a learned state aggregation, and a soft-alignment auxiliary. As a result, \textsc{LoopMTP} obtains up to 8.1\% relative average gains over a parameter-matched non-looped baseline, wins over the state-of-the-art looped model in 27 of 28 matched comparisons, and demonstrates stable training at even large loop counts. We further show \textsc{LoopMTP} can be effectively employed to train small domain-specific expert models. By training a math-expert model, \textsc{LoopMTP} reaches a 11.98 p.p.~improvement on GSM8K over the non-looped baseline. As a next step, we aim to investigate the scaling behaviour of \textsc{LoopMTP} and establish corresponding scaling laws.

\section{Limitations}
The small-expert results in Section~\ref{sec:expert} cover only the math domain; we selected it as a representative task demanding multi-step reasoning, but transfer of these benefits to other specialized domains remains to be verified.
While the main results in Section~\ref{sec:main_results} are averaged over three seeds, with standard deviations reported in Appendix~\ref{sec:appendix:multiseed_results}, the domain-expert models in Section~\ref{sec:expert} were each trained with a single seed, a constraint imposed by our computational budget. However, the consistency of improvements as well as stable performance of \textsc{LoopMTP} in other settings might mitigate concerns about seed sensitivity.
Finally, we observe diminishing and occasionally non-monotonic gains as the loop count grows; a full characterization of the effective upper bound on useful depth for looped architectures remains an open question.

\section{Ethics Statement}

This work proposes a pretraining architecture. It introduces no new datasets,
deploys no system, and involves no human subjects. All models are decoder-only
transformers trained from scratch on publicly released corpora and evaluated on
standard public benchmarks, in accordance with their licenses.

\textsc{LoopMTP} changes how computation is organized within a fixed parameter budget and is agnostic to training data and downstream tasks; it therefore inherits, but does not amplify, the known limitations of transformer language models trained on web text; we do not expect the architectural changes themselves to introduce new categories of risk.
We see a concrete ethical benefit in parameter-efficient models of this kind: by improving parameter efficiency, they can be deployed fully on-premises, allowing sensitive data to remain under the data holder's control rather than being sent to third-party APIs. 

AI assistants were used for copy-editing, LaTeX formatting, and plotting
support. The authors were responsible for all research ideas, experimental design, analysis, and conclusions.

\bibliography{custom}

\clearpage

\appendix
\label{sec:appendix}
\section{Implementation Details}
\label{sec:appendix:implementation_details}

The hyperparameter values for our main experiments are shown in Table~\ref{tab:hyperparams_config}. One choice worth highlighting is the value of $\lambda_{\text{align}}$ for each loop count, which was obtained by a sweep over \{0.01, 0.05, 0.1, 0.15, 0.3, 0.4\} to get the optimal performance. The overall increasing trend indicates that larger loop counts yield gradient contributions from a greater number of iterations, potentially strengthening the effective optimization signal and allowing more aggressive alignment of intermediate representations without sacrificing performance. 

For the training runs in Section~\ref{sec:expert}, all hyperparameters remain the same except for gate initializations and optimizer settings. The first and second gates are initialized to $0.55$ and $-3.0$, with each subsequent iteration decreasing by $0.5$. Learning rate (LR) is set to $1.889{\times}10^{-3}$ and weight decay (WD) to $0.132$. 

For LoopFormer, we perform a grid search over WD, warmup ratio, and LR for each loop configuration. The LR is swept over $\{6{\times}10^{-4}, 1.9{\times}10^{-3}, 3.8{\times}10^{-3}\}$, corresponding to the value reported in the original paper, the value used for \textsc{LoopMTP}, and a higher setting. WD is swept over \{0.1, 0.2\} and warmup ratio over \{0.001, 0.08\}, covering both the \textsc{LoopMTP} values and those reported by \citet{jeddi2026loopformer}. This yields 12 configurations per loop count.
Crucially, LoopFormer's grid search targets \emph{training stability}:
without the right combination
training diverges due to gradient explosion. By contrast, \textsc{LoopMTP}
trains stably across all loop counts under a single fixed
configuration of stability-critical hyperparameters. The only
per-loop adjustment is $\lambda_{\text{align}}$, which controls the
strength of the auxiliary alignment signal: all swept values produce
stable training runs, and the sweep serves solely
downstream performance.

\begin{table}[t]
\centering

\resizebox{\linewidth}{!}{
\begin{tabular}{@{}clr@{}}
\toprule
Category & Parameter & Value \\
\midrule
\multirow{11}{*}{\makecell{\textbf{Model} \\\textbf{Architecture}}}
  & Layers                             & 12 \\
  & Embedding dim                      & 1024 \\
  & Query heads                        & 32 \\
  & KV heads                           & 32 \\
  & FFN hidden dim (nominal $4d$)      & 4096 \\
  & Tokenizer                          & GPT-2 \cite{radford2019language}\\
  & Vocabulary size                    & 50304 \\
  & Position encoding                  & RoPE \\
  & Activation                         & SwiGLU \\
  & Normalization layer                & RMSNorm \\
  & Weight tying                       & \ding{55} \\
\midrule
\multirow{5}{*}{\makecell{\textbf{Looping}}}
  & Iterations ($T$)                  & \{3,5,7,9\} \\
  & Gates initial bias                 & First:$0.55$ , Rest:$-3.0$ \\
  & Shared gate                        & \cmark \\
  & Per-iter. norms (hidden)           & \cmark \\
  & Per-iter. norms (token embd)       & \ding{55} \\
\midrule
\multirow{3}{*}{\textbf{Loss}}
      
  & $\lambda_{\text{ponder}}$          & 0.05 \\
  & $\lambda_{\text{align}}$           & \{0.01, 0.01, 0.05, 0.15\} \\
  & Alignment type                     & Cosine \\
\midrule
\multirow{6}{*}{\textbf{Optimizer}}
  & Optimizer                          & Muon \\
  & Learning rate                      & $1.9 \times 10^{-3}$ \\
  & AdamW $\beta$                      & $(0.9,\,0.95)$ \\
  & Weight decay (WD)                  & 0.1 \\
  & WD excluded layers                 & Embedding, Layer/RMSnorm\\
  & Gradient clip                      & 1.0 \\
\midrule
\multirow{3}{*}{\makecell{\textbf{LR} \\\textbf{Scheduler}}}
  & Warmup ratio                       & 0.001 \\
  & Div factor                         & 10 \\
  & Anneal strategy                    & Cosine \\
\midrule
\multirow{5}{*}{\textbf{Training}}
  & Mixed precision                    & BF16 \\
  & Sequence length                    & 2048 \\
  & Optimization steps                 & 27,864 \\
  & Effective batch size               & $\approx$245K \\
  & Total training tokens              & $\approx 6.8$B \\
\bottomrule
\end{tabular}
}
\caption{\textbf{Key design decisions and hyperparameter values.} The values used for \textsc{LoopMTP} in Section~\ref{sec:results}.}
\label{tab:hyperparams_config}
\end{table}

\definecolor{archcolor}{HTML}{EBF2FF}   %
\definecolor{loopcolor}{HTML}{FFF0E0}   %
\definecolor{losscolor}{HTML}{E6FAF0}   %
\definecolor{optcolor}{HTML}{FDF0F4}    %
\definecolor{lrcolor}{HTML}{FDF0F4}     %
\definecolor{traincolor}{HTML}{FFF8E1}  %

\section{An Observation on Loop Count and Reasoning-Step Length} 
\label{sec:appen:mechanistic_verification}

\begin{figure}[t]
    \centering
    \includegraphics[
        width=\linewidth,
        trim=0cm 9pt 0cm 12pt, %
        clip
    ]{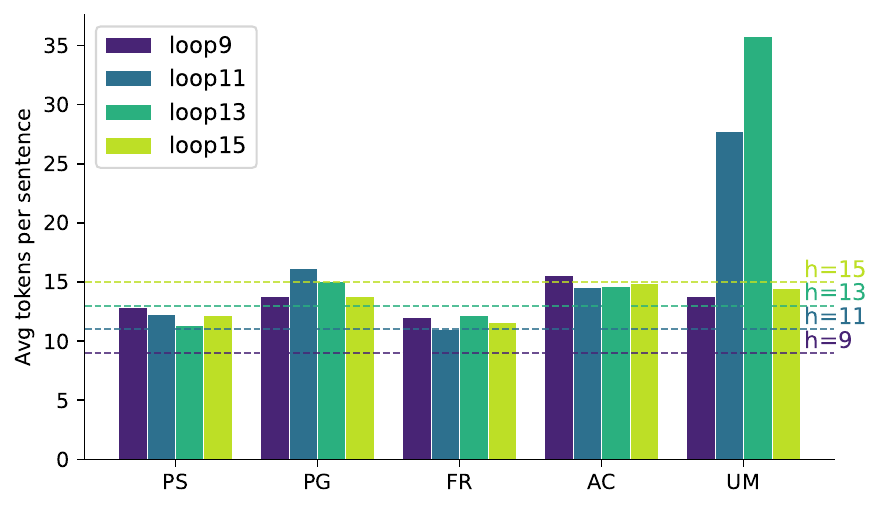}
    \vspace{-20pt}
    \caption{\textbf{Average number of tokens per annotated tag}. The reasoning traces are annotated following \citet{bogdan2025thought}.
    }
    \vspace{-10pt}
    \label{fig:avg_tokens_per_tag}
\end{figure}

To generate hypotheses about why certain loop counts outperform others
(e.g., 9 vs.\ 11), we conduct an exploratory analysis of GSM8K responses
from Section~\ref{sec:expert}. We stress that the evidence below is
\emph{correlational}. We use MiniMax-M2.5 \citep{chen2026minimax} to
annotate responses following \citet{bogdan2025thought} and measure average
tokens per sentence within each semantic tag. The resulting distributions
(Figure~\ref{fig:avg_tokens_per_tag}) suggest a coupling between MTP
lookahead depth and the natural token length of reasoning steps, which we
term the \emph{horizon alignment} conjecture.

\paragraph{Canonical sentence lengths define the alignment target.}
Across loop configurations, the model's reasoning vocabulary settles into a
narrow band of characteristic lengths: \texttt{active\_computation} (AC)
steps average ${\sim}14.5$--$15.5$ tokens, while
\texttt{problem\_setup} (PS) and \texttt{fact\_retrieval} (FR) steps cluster
at ${\sim}11$--$13$ tokens, and \texttt{plan\_generation} (PG) steps average ${\sim}14$--$16$ tokens. These values vary little with loop depth,
marking them as intrinsic structural units of the model's reasoning rather
than artifacts of a particular decoding configuration.

\paragraph{A possible account of the \texorpdfstring{$\text{Loop}15$}{loop 15} peak.} A 15-token lookahead is the smallest window that fully contains a
canonical AC step from opening clause to final result. Because the model can
preview an entire computational thought within a single horizon, it never
commits to an intermediate token without visibility into where the
calculation lands, explaining the accuracy peak at $\text{Loop}15$
($19.03\%$).

\begin{figure*}[t]
    \centering
    \includegraphics[
        width=\linewidth,
        trim=0cm 7pt 0cm 6pt, %
        clip
    ]{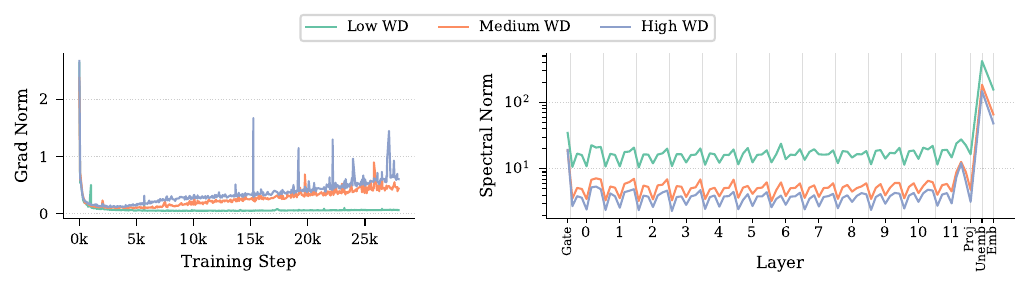}
    \vspace{-7pt}
    \caption{\textbf{Effect of WD on a looped model.} 
    (Left) Gradients during training for $T{=}9$ model
    (Right) Per-layer spectral norms of the trained model. 
    }
    \label{fig:spectral_norm}
\end{figure*}

\paragraph{An unexplained dip at \texorpdfstring{$\text{Loop}11$}{loop 11} and \texorpdfstring{$\text{Loop}13$}{loop 13}.} 
When the lookahead window neither spans a full execution step nor coincides with a stable
planning boundary, performance degrades sharply. At $\text{Loop}11$, the
horizon falls just short of the ${\sim}14.5$ token AC baseline; unable to preview the
outcome of a computation already begun, the model enters a recovery mode
visible as bloated \texttt{uncertainty\_management} (UM) sentences averaging
${\sim}28$ tokens---verbose, structurally unanchored attempts to re-derive
context the window could not supply. At $\text{Loop}13$, the horizon is wide
enough to \emph{enter} an AC step but too narrow to \emph{exit} it cleanly,
trapping generation in mid-computation limbo (${\sim}35$ UM tokens). Crucially, misalignment does
not suppress generation but redirects it into unproductive verbosity.

We stress the limits of this account. It offers a plausible reading of the $\text{Loop}15$ peak and the $\text{Loop}11$ / $\text{Loop}13$ dip, but it does not explain the strong results at $\text{Loop}7$ (18.12\%) and $\text{Loop}9$ (18.73\%), whose lookahead horizons are further from the canonical AC length than those of the configurations that underperform. A mechanism operating solely through horizon–step alignment would predict the opposite ordering for these points.

\section{Importance of Weight Decay Value for Looped Models}
\label{sec:appendix:WD}

While weight decay (WD) is standard in language model training
\citep{loshchilov2017decoupled}, it is typically left at its default value.
Our experiments reveal that WD has a surprisingly large impact on reasoning
performance for looped models. To isolate this effect, we fix
$\lambda_\text{align}{=}0.3$ throughout this section.

As shown in Figure~\ref{fig:spectral_norm} (Left), low WD leads to rapidly diminishing gradient norms over training, while medium and high WD sustain larger norms throughout. We note that the observed pattern is consistent with the view that weight decay acts less as a classical regularizer than as a modifier of optimization dynamics \citep{d2024we}. We hypothesize this effect is
amplified in looped architectures: as the function is composed $T$ times,
the sensitivity introduced by sharp minima compounds across loops.

\begin{table}[t]
  \centering
    \centering
    \setlength{\tabcolsep}{3.1pt}
    \resizebox{\linewidth}{!}{
    \begin{tabular}{l ccccc}
        \toprule
        & \multicolumn{4}{c}{GSM8K Accuracy $\uparrow$} \\ 
        \cmidrule(lr){2-5}
         WD &  Non-looped & $T{=}3$ & $T{=}5$ & $T{=}9$\\ \midrule
         Low (0.0132)   & 5.00              & 8.42              & 11.44             & 13.57\\
         Medium (0.132) & \textbf{7.05}     & \textbf{12.88}    & \underline{14.70} & \textbf{18.49}\\
         High (0.2)     & \underline{5.07}  & \underline{12.13} & \textbf{16.98}    & \underline{18.12}\\
         \bottomrule
    \end{tabular}
    }
    \caption{\textbf{Performance of looped models for different weight decay values.} Weight decay (WD) alone accounts for relative differences of up to 53\% between the best- and worst-performing settings at a fixed number of iterations.}
    \vspace{-10pt}
    \label{tab:wd_comparison}
\end{table}

Higher WD also constrains weight norms
(Figure~\ref{fig:spectral_norm}, Right), reducing layer Lipschitz constants
\citep{bartlett2017spectrally} and improving stability under repeated
application. On GSM8K (Table~\ref{tab:wd_comparison}), medium and high WD
consistently outperform low WD in both looped and non-looped settings, with
up to 53\% relative improvement. High WD peaks at $T{=}5$ (16.98\% vs.\
11.44\%) but is surpassed by medium WD at $T{=}9$ (18.12\% vs.\ 18.49\%),
suggesting excessive decay over-constrains capacity. These results identify
WD as a surprisingly impactful yet undertuned hyperparameter for looped
models.

\section{FLOPs Comparison}
\label{sec:flops}

\begin{table*}[t]
\centering
\setlength{\tabcolsep}{6pt}
\newcommand{\rel}[1]{{\scriptsize\,(#1$\times$)}}
\resizebox{0.8\linewidth}{!}{
\begin{tabular}{@{}lcccc@{\hskip 14pt}ccc@{}}
\toprule
& \multicolumn{3}{c}{Inference (FLOPs/token, $\times10^{9}$)}
& \multicolumn{3}{c}{Training (total FLOPs, $\times10^{19}$)} \\
\cmidrule(r){2-4}\cmidrule(l){5-7}
$T$ & \textsc{LoopMTP} & LoopFormer & $\Delta$
    & \textsc{LoopMTP} & LoopFormer & $\Delta$ \\
\midrule
Non-looped & \multicolumn{2}{c}{0.481} & --
           & \multicolumn{2}{c}{0.988} & -- \\
\midrule
3 & 1.20\rel{2.49} & 1.31\rel{2.73} & $-8.8\%$ & 2.46\rel{2.49} & 4.15\rel{4.20} & $-40.7\%$ \\
5 & 1.93\rel{4.01} & 2.12\rel{4.40} & $-8.9\%$ & 3.96\rel{4.01} & 6.63\rel{6.71} & $-40.2\%$ \\
7 & 2.66\rel{5.53} & 2.92\rel{6.08} & $-9.0\%$ & 5.47\rel{5.53} & 9.11\rel{9.23} & $-40.1\%$ \\
9 & 3.39\rel{7.05} & 3.73\rel{7.75} & $-9.0\%$ & 6.97\rel{7.05} & 11.6\rel{11.74} & $-39.9\%$ \\
\bottomrule
\end{tabular}
}
\caption{\textbf{FLOPs comparison between \textsc{LoopMTP}, its non-looped baseline and LoopFormer \cite{jeddi2026loopformer}.} Parenthesised values are multiples of the
non-looped baseline. $\Delta$ is the relative change of \textsc{LoopMTP}
with respect to LoopFormer; negative values indicate that
\textsc{LoopMTP} requires less computation.}
\label{tab:flops-loopmtp-vs-loopformer}
\end{table*}

Following the accounting of the LoopFormer \cite{jeddi2026loopformer}, we write the cost of a
forward pass with $t$ loop iterations as $C(t) = C_{io} + t\, C_1$, where
$C_{io}$ collects the loop-independent input/output compute (final norm + unembedding;
embedding lookups are free) and $C_1$ is the cost of one pass through the shared
$L$-block stack plus each model's per-iteration machinery. 
The non-looped baseline is the $T=1$ instance of the same expression, with a slightly wider \textsc{SwiGLU}. \textsc{LoopMTP} follows the LLaMA sizing rule ($\tfrac{2}{3}\cdot 4d$, rounded up to the nearest multiple of $256$), so the nominal $4096$ of Table~\ref{tab:hyperparams_config} corresponds to a realized hidden width of $2816$ at $d{=}1024$. For the baseline, which has no input projection or aggregation gate, we widen this to the next multiple of $256$, i.e.\ $3072$. Because the width is constrained to multiples of $256$, this does not match the recurrence parameters exactly: it overshoots them, leaving the baseline roughly $6$M parameters \emph{larger} than \textsc{LoopMTP}.
We count $2$ FLOPs per multiply--accumulate, score attention
causally, and take one training step to cost $3\times$ the forward pass.
For our configuration ($L=12$, $d=1024$, $S=2048$, $|V|=50304$, $D \approx 6.8$B), per token:
\begin{align*}
\text{Baseline:}\quad &
\left\{
\begin{aligned}
    C_{io}/S &= 1.03 \times 10^{8}\\ 
    C_1^{\mathrm{base}}/S &= 3.78 \times 10^{8}
\end{aligned}
\right. \\[7pt]
\text{LoopMTP:}\quad &
\left\{
\begin{aligned}
    C_{io}/S &= 1.03 \times 10^{8}\\
    C_1/S &= 3.65 \times 10^{8}
\end{aligned}
\right. \\[7pt]
\text{LoopFormer:}\quad &
\left\{
\begin{aligned}
    C_{io}/S &= 1.03 \times 10^{8}\\
    C_1/S &= 4.03 \times 10^{8}
\end{aligned}
\right.
\end{align*}

Inference uses a single $T$-loop trajectory for both models,
$C_{\mathrm{inf}}(T) = C_{io} + T\,C_1$ and a single pass for the baseline,
$C_{\mathrm{inf}}^{\mathrm{base}} = C_{io} + C_1^{\mathrm{base}}$ . Over a token budget $D$, training costs:
\begin{align*}
\mathcal{C}^{\mathrm{train}}_{\mathrm{base}} &= 3\,\big(C_{io} + C_1^{\mathrm{base}}\big)\, D/S, \\[8pt]
\mathcal{C}^{\mathrm{train}}_{\mathrm{LoopMTP}} &= 3\,\big(C_{io} + T\, C_1\big)\, D/S, \\[8pt]
\mathbb{E}\big[\mathcal{C}^{\mathrm{train}}_{\mathrm{LoopFormer}}\big] &= 3\,\big(2\,C_{io} + \tfrac{3T}{2}\, C_1\big)\, D/S,
\end{align*}
since LoopFormer backpropagates both its full $T$-loop trajectory and a short
$M$-loop trajectory with $M \sim \mathrm{Unif}\{1,\dots,T-1\}$ ($\mathbb{E}[M] = T/2$).

Table~\ref{tab:flops-loopmtp-vs-loopformer} compares the training and inference FLOPs of LoopFormer and \textsc{LoopMTP}. For a matched loop count $T$ and token budget, \textsc{LoopMTP} requires only 0.59--0.60$\times$ the training FLOPs of LoopFormer, while achieving better performance. The additional computational cost of LoopFormer primarily stems from its dual-trajectory objective. During inference, \textsc{LoopMTP} remains approximately 10\% more FLOP-efficient.
Relative to the non-looped baseline of the same size, \textsc{LoopMTP} costs 2.49--7.05$\times$ the FLOPs at both training and inference for $T{=}3$--$9$. Importantly, looping converts compute into effective depth; it does not reduce total computation relative to a shallow model.

\section{Detailed Result}
\label{sec:appendix:multiseed_results}

In this section, we present more detailed results, including a breakdown of the aggregated numbers reported in Table~\ref{tab:main_results}, along with the average and standard deviation (STD) values. All results in this section report the average performance for each configuration across 3 random seeds (unless otherwise noted), together with the corresponding standard deviation.

\definecolor{lightblue}{RGB}{220,230,255}
\definecolor{lightgray}{RGB}{240,240,240}

\begin{table*}[t]
    \centering
    \setlength{\tabcolsep}{4.5pt}
    \resizebox{0.85\linewidth}{!}{
    \begin{tabular}{
        l |
        c c c c c c c c
        }
    \toprule
        & \multicolumn{8}{c}{\textit{General \& QA Tasks (Accuracy $\uparrow$)}}
        \\
        \cmidrule(lr){2-9}
        Model
        & ARC-C & ARC-E & HS & WG & SIQA & PIQA & LB & \textit{Avg}
        \\ \midrule
        \textbf{Non-looped} & 31.60$_{1.16}$ & 59.90$_{0.54}$ & 38.48$_{0.04}$ & 50.96$_{0.27}$ & 44.85$_{0.27}$ & 65.98$_{0.24}$ & 32.16$_{0.89}$ & 46.28$_{0.28}$
        \\ \midrule
        
        \textbf{\textit{LoopFormer$^*$}}
        \\
        \quad Loops=3$^\dagger$ & 32.04$_{0.81}$ & 58.69$_{2.04}$ & 38.61$_{2.56}$ & 50.24$_{0.67}$ & 44.19$_{1.97}$ & 66.00$_{0.98}$ & 30.98$_{3.29}$ & 45.82$_{1.76}$\\
        \quad Loops=5 & 35.24$_{1.40}$ & 59.65$_{0.98}$ & 41.61$_{0.86}$ & 52.35$_{0.84}$ & 45.87$_{0.23}$ & 65.29$_{0.56}$ & 35.13$_{0.62}$ & 47.88$_{0.49}$\\ 
        \quad Loops=7 & 30.01$_{3.44}$ & 46.49$_{14.34}$ & 34.37$_{7.30}$ & 51.43$_{1.55}$ & 42.39$_{3.66}$ & 60.52$_{7.25}$ & 20.84$_{15.24}$ & 40.86$_{7.48}$\\ 
        \quad Loops=9 & 32.17$_{3.20}$ & 49.97$_{16.38}$ & 36.50$_{7.66}$ & 50.86$_{1.68}$ & 42.94$_{3.98}$ & 60.97$_{7.96}$ & 23.73$_{16.78}$ & 42.45$_{8.22}$\\ \midrule
        
        \rowcolor{lightgray} \textit{\textbf{\textsc{LoopMTP}}\small(ours)\normalsize} &&&&&&&&\\
        \rowcolor{lightgray} \quad Loops=3 & 35.49$_{1.24}$ & 62.28$_{0.93}$ & 42.60$_{0.32}$ & 50.07$_{0.71}$ & 46.37$_{0.81}$ & 67.17$_{0.13}$ & 35.48$_{1.08}$ & 48.49$_{0.08}$\\
        \rowcolor{lightgray} \quad Loops=5 & 35.32$_{1.79}$ & 63.38$_{0.72}$ & 44.35$_{0.35}$ & 52.54$_{0.46}$ & 46.88$_{0.59}$ & 67.52$_{0.24}$ & 36.79$_{1.09}$ & 49.54$_{0.36}$\\
        \rowcolor{lightgray} \quad Loops=7 & 35.10$_{0.89}$ & \textbf{64.07$_{0.69}$} & \textbf{44.78$_{0.29}$} & \textbf{52.99$_{0.83}$} & \underline{46.95}$_{0.51}$ & \textbf{67.97$_{0.85}$} & 36.39$_{1.44}$ & \underline{49.75}$_{0.31}$\\
        \rowcolor{lightgray} \quad Loops=9 & \textbf{36.01$_{0.39}$} & \underline{63.71}$_{0.53}$ & \underline{44.64}$_{0.16}$ & \underline{52.70}$_{0.35}$ & \textbf{47.78$_{0.16}$} & \underline{67.86}$_{0.33}$ & \textbf{37.47$_{1.46}$} & \textbf{50.02$_{0.16}$}\\
    \bottomrule
    \end{tabular}
    }
    \caption{\textbf{Accuracy results on general and QA tasks, with multi-seed standard deviations.}
    $^*$ for LoopFormer \cite{jeddi2026loopformer}, stable training required a separate grid search over learning rate, weight decay, and warmup ratio at each loop count. $^\dagger$ indicates runs whose reported averages are over two random seeds, as one of the three runs diverged.}
    \label{tab:acc_general_results}
\end{table*}

\begin{table*}[t]
    \centering
    \setlength{\tabcolsep}{4.5pt}
    \resizebox{\linewidth}{!}{
    \begin{tabular}{
        l |
        c c c c c c c c
        }
    \toprule
        & \multicolumn{8}{c}{\textit{General \& QA Tasks (BPB $\downarrow$)}}
        \\
        \cmidrule(lr){2-9}
        Model
        & ARC-C & ARC-E & HS & WG & SIQA & PIQA & LB & \textit{Avg}
        \\ \midrule
        \textbf{Non-looped} & 0.8842$_{0.0008}$ & 1.1489$_{0.0030}$ & 0.8280$_{0.0106}$ & 0.7493$_{0.0065}$ & 0.9543$_{0.0110}$ & 1.2262$_{0.0020}$ & 1.1723$_{0.0061}$ & 0.9948$_{0.0045}$
        \\ \midrule
        
        \textbf{\textit{LoopFormer$^*$}}
        \\
        \quad Loops=3$^\dagger$ & 0.9017$_{0.0199}$ & 1.1625$_{0.0211}$ & 0.8637$_{0.0665}$ & 0.7815$_{0.0688}$ & 0.9782$_{0.0587}$ & 1.2500$_{0.0262}$ & 1.2109$_{0.0274}$ & 1.0212$_{0.0412}$\\
        \quad Loops=5 & 0.8771$_{0.0049}$ & 1.1494$_{0.0125}$ & 0.7895$_{0.0113}$ & 0.7134$_{0.0120}$ & 0.9086$_{0.0134}$ & 1.2196$_{0.0043}$ & 1.1655$_{0.0124}$ & 0.9747$_{0.0083}$\\ 
        \quad Loops=7 & 1.3658$_{0.6304}$ & 1.6384$_{0.6571}$ & 1.6467$_{1.0783}$ & 1.2893$_{0.7270}$ & 1.4615$_{0.6869}$ & 1.6241$_{0.5363}$ & 1.5720$_{0.5020}$ & 1.5140$_{0.6882}$\\ 
        \quad Loops=9 & 1.3005$_{0.6070}$ & 1.5730$_{0.6345}$ & 1.5596$_{1.1166}$ & 1.2559$_{0.8001}$ & 1.3973$_{0.7121}$ & 1.5852$_{0.5335}$ & 1.5168$_{0.4966}$ & 1.4555$_{0.7000}$\\ \midrule
        
        \rowcolor{lightgray} \textit{\textbf{\textsc{LoopMTP}}\small(ours)\normalsize} &&&&&&&&\\
        \rowcolor{lightgray} \quad Loops=3 & 0.8600$_{0.0004}$ & 1.1155$_{0.0057}$ & 0.7768$_{0.0171}$ & 0.6885$_{0.0018}$ & 0.8943$_{0.0058}$ & 1.2005$_{0.0085}$ & 1.1431$_{0.0051}$ & 0.9541$_{0.0013}$\\
        \rowcolor{lightgray} \quad Loops=5 & 0.8551$_{0.0024}$ & 1.1129$_{0.0050}$ & 0.7492$_{0.0064}$ & 0.6666$_{0.0028}$ & \textbf{0.8721$_{0.0077}$} & \textbf{1.1913$_{0.0119}$} & 1.1421$_{0.0046}$ & 0.9413$_{0.0033}$\\
        \rowcolor{lightgray} \quad Loops=7 & \textbf{0.8515$_{0.0033}$} & \underline{1.1121}$_{0.0030}$ & \underline{0.7473}$_{0.0142}$ & \textbf{0.6626}$_{0.0019}$ & \underline{0.8730}$_{0.0036}$ & \underline{1.1965}$_{0.0025}$ & \textbf{1.1151$_{0.0102}$} & \underline{0.9369}$_{0.0022}$\\
        \rowcolor{lightgray} \quad Loops=9 & \underline{0.8526}$_{0.0022}$ & \textbf{1.1086$_{0.0068}$} & \textbf{0.7359$_{0.0220}$} & \underline{0.6649}$_{0.0070}$ & 0.8756$_{0.0027}$ & 1.1980$_{0.0152}$ & \underline{1.1204}$_{0.0107}$ & \textbf{0.9366$_{0.0050}$}\\
    \bottomrule
    \end{tabular}
    }
    \caption{\textbf{BPB results on general and QA tasks, with multi-seed standard deviations.}
    $^*$ for LoopFormer \cite{jeddi2026loopformer}, stable training required a separate grid search over learning rate, weight decay, and warmup ratio at each loop count. $^\dagger$ indicates runs whose reported averages are over two random seeds, as one of the three runs diverged.}
    \label{tab:bpb_general_results}
\end{table*}

\begin{table*}[t]
    \centering
    \setlength{\tabcolsep}{4.5pt}
    \resizebox{\linewidth}{!}{
    \begin{tabular}{
        l |
        c c c c c c c c
        }
    \toprule
        & \multicolumn{8}{c}{\textit{Math \& Reasoning Tasks (BPB $\downarrow$)}}
        \\
        \cmidrule(lr){2-9}
        Model
        & algebra & \makecell{counting \&\\probability} & geometry & \makecell{intermediate\\algebra} & \makecell{number\\theory} & prealgebra & precalculus & \textit{Avg}
        \\ \midrule
        \textbf{Non-looped} & 0.7015$_{0.0029}$ & 0.6839$_{0.0023}$ & 0.7696$_{0.0032}$ & 0.7321$_{0.0047}$ & 0.7661$_{0.0009}$ & 0.6696$_{0.0019}$ & 0.6572$_{0.0028}$ & 0.7114$_{0.0024}$
        \\ \midrule
        
        \textbf{\textit{LoopFormer$^*$}}
        \\
        \quad Loops=3$^\dagger$ & 0.7260$_{0.0442}$ & 0.7014$_{0.0391}$ & 0.7963$_{0.0472}$ & 0.7579$_{0.0445}$ & 0.7889$_{0.0384}$ & 0.6901$_{0.0382}$ & 0.6801$_{0.0433}$ & 0.7344$_{0.0421}$\\
        \quad Loops=5 & 0.6789$_{0.0133}$ & 0.6604$_{0.0113}$ & 0.7543$_{0.0095}$ & 0.7053$_{0.0116}$ & 0.7461$_{0.0127}$ & 0.6490$_{0.0129}$ & 0.6348$_{0.0108}$ & 0.6898$_{0.0116}$\\ 
        \quad Loops=7 & 1.7655$_{1.4466}$ & 1.5079$_{1.1230}$ & 1.8546$_{1.4597}$ & 1.8537$_{1.5247}$ & 1.7187$_{1.2894}$ & 1.5966$_{1.2564}$ & 1.7844$_{1.5343}$ & 1.7259$_{1.3763}$\\ 
        \quad Loops=9 & 1.6296$_{1.3613}$ & 1.4130$_{1.0741}$ & 1.7195$_{1.3811}$ & 1.7134$_{1.4341}$ & 1.6130$_{1.2377}$ & 1.4812$_{1.1884}$ & 1.6545$_{1.4508}$ & 1.6034$_{1.3039}$\\ \midrule
        
        \rowcolor{lightgray} \textit{\textbf{\textsc{LoopMTP}}\small(ours)\normalsize} &&&&&&&&\\
        \rowcolor{lightgray} \quad Loops=3 & 0.6526$_{0.0020}$ & 0.6442$_{0.0034}$ & 0.7302$_{0.0029}$ & 0.6833$_{0.0032}$ & 0.7240$_{0.0016}$ & 0.6289$_{0.0019}$ & 0.6138$_{0.0031}$ & 0.6681$_{0.0022}$\\
        \rowcolor{lightgray} \quad Loops=5 & \textbf{0.6410$_{0.0022}$} & 0.6331$_{0.0010}$ & 0.7190$_{0.0031}$ & \underline{0.6784}$_{0.0008}$ & \underline{0.7130}$_{0.0019}$ & 0.6169$_{0.0005}$ & \textbf{0.6077$_{0.0029}$} & \underline{0.6584}$_{0.0014}$\\
        \rowcolor{lightgray} \quad Loops=7 & \underline{0.6414}$_{0.0030}$ & \textbf{0.6306$_{0.0017}$} & \textbf{0.7179$_{0.0013}$} & \textbf{0.6771$_{0.0035}$} & \textbf{0.7100$_{0.0023}$} & \underline{0.6153}$_{0.0025}$ & \underline{0.6101}$_{0.0021}$ & \textbf{0.6575$_{0.0021}$}\\
        \rowcolor{lightgray} \quad Loops=9 & \textbf{0.6410$_{0.0040}$} & \underline{0.6322}$_{0.0024}$ & \underline{0.7180}$_{0.0061}$ & 0.6819$_{0.0073}$ & 0.7132$_{0.0045}$ & \textbf{0.6137$_{0.0032}$} & 0.6103$_{0.0071}$ & 0.6586$_{0.0049}$\\
    \bottomrule
    \end{tabular}
    }
    \caption{\textbf{BPB results on math and reasoning tasks, with multi-seed standard deviations.}
    $^*$ for LoopFormer \cite{jeddi2026loopformer}, stable training required a separate grid search over learning rate, weight decay, and warmup ratio at each loop count. $^\dagger$ indicates runs whose reported averages are over two random seeds, as one of the three runs diverged.}
    \label{tab:bpb_math_results}
\end{table*}

\begin{table*}[t]
    \centering
    \setlength{\tabcolsep}{4.5pt}
    \resizebox{0.59\linewidth}{!}{
    \begin{tabular}{
        l |
        c c c
        }
    \toprule
        & \multicolumn{3}{c}{\textit{Coding Tasks (BPB $\downarrow$)}}
        \\
        \cmidrule(lr){2-4}
        Model
        & mbpp & \makecell{codex\\humaneval}
        & \textit{Avg}
        \\ \midrule
        \textbf{Non-looped} & 0.9710$_{0.0033}$ & 0.7538$_{0.0048}$ & 0.8624$_{0.0040}$
        \\ \midrule
        
        \textbf{\textit{LoopFormer$^*$}}
        \\
        \quad Loops=3$^\dagger$ & 0.9922$_{0.0506}$ & 0.8152$_{0.0726}$ & 0.9037$_{0.0616}$\\
        \quad Loops=5 & 0.9320$_{0.0153}$ & 0.7375$_{0.0186}$ & 0.8348$_{0.0163}$\\ 
        \quad Loops=7 & 2.4820$_{2.0270}$ & 2.1205$_{1.7762}$ & 2.3012$_{1.9015}$\\ 
        \quad Loops=9 & 1.9664$_{1.4739}$ & 1.6515$_{1.3134}$ & 1.8089$_{1.3937}$\\ \midrule
        
        \rowcolor{lightgray} \textit{\textbf{\textsc{LoopMTP}}\small(ours)\normalsize} &&&\\
        \rowcolor{lightgray} \quad Loops=3 & 0.9072$_{0.0113}$ & 0.7070$_{0.0107}$ & 0.8071$_{0.0083}$\\
        \rowcolor{lightgray} \quad Loops=5 & \underline{0.8876}$_{0.0042}$ & \underline{0.6965}$_{0.0062}$ & \underline{0.7921}$_{0.0037}$\\
        \rowcolor{lightgray} \quad Loops=7 & \textbf{0.8812$_{0.0053}$} & \textbf{0.6832$_{0.0017}$} & 0.7822$_{0.0019}$\\
        \rowcolor{lightgray} \quad Loops=9 & 0.8942$_{0.0115}$ & 0.7064$_{0.0087}$ & 0.8003$_{0.0101}$\\
    \bottomrule
    \end{tabular}
    }
    \caption{\textbf{BPB results on coding tasks, with multi-seed standard deviations.}
    $^*$ for LoopFormer \cite{jeddi2026loopformer}, stable training required a separate grid search over learning rate, weight decay, and warmup ratio at each loop count. $^\dagger$ indicates runs whose reported averages are over two random seeds, as one of the three runs diverged.}
    \label{tab:bpb_code_results}
\end{table*}

Table~\ref{tab:acc_general_results} shows the accuracy for general and QA tasks. Here, \textsc{LoopMTP} not only consistently outperforms the non-looped baseline and LoopFormer (27 out of 28 settings; the exception is Winogrande at $T$=3), but also shows more stable performance, i.e.\ lower STD values. While the STD values for \textsc{LoopMTP} consistently remain in a very low range, LoopFormer's performance suffers from large oscillations (despite its hyperparameters being tuned separately for each loop count): for 3-loop training, one of LoopFormer's runs diverged (which is why the 3-loop LoopFormer results are aggregated over only two runs), and for 7 and 9 loops, one of the runs did not perform very well after training. Tables~\ref{tab:bpb_general_results}--\ref{tab:bpb_code_results} show the BPB values separately for each aggregated benchmark. The same pattern holds: \textsc{LoopMTP} consistently outperforms both the non-looped and LoopFormer baselines.

\end{document}